\documentclass{article}

\usepackage{microtype}
\usepackage{graphicx}
\usepackage{subfigure}
\usepackage{booktabs} 
\usepackage{caption}
\usepackage{comment}

\usepackage{hyperref}

\usepackage[accepted]{icml2025}

\usepackage{amsmath}
\usepackage{amssymb}
\usepackage{mathtools}
\usepackage{amsthm}

\usepackage[capitalize,noabbrev]{cleveref}

\theoremstyle{plain}

\theoremstyle{definition}

\theoremstyle{remark}

\usepackage[textsize=tiny]{todonotes}

\icmltitlerunning{TabPFN Through The Looking Glass}

\begin{document}

\twocolumn[
\icmltitle{TabPFN Through The Looking Glass: \\
           An interpretability study of TabPFN and its internal representations}



\icmlsetsymbol{equal}{*}

\begin{icmlauthorlist}
\icmlauthor{Aviral Gupta}{sch}
\icmlauthor{Armaan Sethi}{sch}
\icmlauthor{Dhruv Kumar}{sch}
\end{icmlauthorlist}

\icmlaffiliation{sch}{Birla Institute of Technology and Science, Pilani}

\icmlcorrespondingauthor{Aviral Gupta}{f20220097@pilani.bits-pilani.ac.in}

\icmlkeywords{Machine Learning, ICML}

\vskip 0.3in
]



\printAffiliationsAndNotice{}  

\begin{abstract}
Tabular foundational models are pre-trained models designed for a wide range of tabular data tasks. They have shown strong performance across domains, yet their internal representations and learned concepts remain poorly understood. This lack of interpretability makes it important to study how these models process and transform input features. In this work, we analyze the information encoded inside the model's hidden representations and examine how these representations evolve across layers. We run a set of probing experiments that test for the presence of linear regression coefficients, intermediate values from complex expressions, and the final answer in early layers. These experiments allow us to reason about the computations the model performs internally. Our results provide evidence that meaningful and structured information is stored inside the representations of tabular foundational models. We observe clear signals that correspond to both intermediate and final quantities involved in the model's prediction process. This gives insight into how the model refines its inputs and how the final output emerges. Our findings contribute to a deeper understanding of the internal mechanics of tabular foundational models. They show that these models encode concrete and interpretable information, which moves us closer to making their decision processes more transparent and trustworthy.
\end{abstract}

\section{Introduction}
\label{introduction}
Tabular Foundational Models are becoming increasingly popular as a tool to replace traditional machine learning techniques in real world prediction tasks. These models are trained at scale on large collections of synthetic datasets generated from structural causal models \citep{causalinferencelearningbook,pearl_2009} and related generative processes. This broad pretraining gives them strong generalization across many tabular distributions and has led to performance that often surpasses classical methods such as gradient boosted trees or shallow neural networks. As a result they are now seen as potential drop in replacements for a wide range of supervised pipelines in domains such as finance \citep{fin_usecase1_nonlife_transformers_actuarial, fin_usecase2_crosssell_health_insurance, fin_usecase3_recovery_rate_github}, healthcare \citep{hc_usecase2_mcd_scirep, hc_usecase3_immunotypes_cancercell,hc_usecase4_stillbirth_slas,hc_usecase5_acc_asj}, manufacturing and industrial engineering \citep{manuf_usecase1_rotating_faults_tabpfn, manuf_usecase2_mcu_performance_tabpfn, manuf_usecase3_caisson_inclination_ml} and scientific data analysis \citep{energy_usecase6_ash_fusibility_high_alkali, energy_usecase7_henry_zeolites, energy_usecase8_shape_selectivity_zeolites}.

However, the growing use of such models introduces a significant challenge. The black box transformer based nature of their architecture makes their processing opaque. Their predictions are produced through stacked attention layers and MLP blocks that obscure the intermediate steps of the computation. This opacity stands in contrast to traditional tabular models, which usually provide clearer reasoning paths or explicit feature contributions. When TabFMs begin replacing these more transparent pipelines, the lack of interpretability becomes a central concern for trust, debugging, and compliance with regulatory requirements. Understanding what information TabFMs store internally and how they compute their outputs becomes increasingly important as these models move toward deployment.

Although TabPFN and related models achieve strong accuracy across a wide range of tasks, there is still very limited understanding of how these architectures represent tabular structure or carry out the computations required for prediction. Existing interpretability work on tabular models mostly focuses on post hoc feature importance, which does not reveal the internal mechanisms that produce the final decision \cite{Rundel_2024}. Meanwhile, the interpretability community studying large language models \citep{zhao2023explainabilitylargelanguagemodels} has shown that transformer representations often contain linearly readable information about tasks \citep{todd2024functionvectorslargelanguage}, intermediate computations, and concepts \citep{park2024linearrepresentationhypothesisgeometry}. These results suggest that transformers learn structured internal algorithms rather than acting as unstructured black box approximators. Since TabPFN relies heavily on in context learning to approximate the target function implied by the provided training data, it becomes natural to ask whether similar algorithmic structure appears inside its hidden representations.

In this work we investigate this question through a set of targeted probing experiments. We construct synthetic datasets where the true functional relationships are known, and we examine how TabPFN v2 \citep{ye2025closerlooktabpfnv2} encodes these relationships inside it's activations. Our probes target three categories of information. First, we test whether the coefficients of simple linear relationships are recoverable from the hidden states. Second, we probe for intermediate quantities that arise inside multi step arithmetic expressions, which allows us to study whether the model represents internal computational steps. Third, we analyze how the final predicted value forms throughout the network using both linear probes \citep{probes} and the logit lens \citep{nostalgebraist2020interpreting} to track the evolution of the output across layers.

Across the experiments we find interpretable structure in the residual stream. Linear coefficients and intermediate arithmetic terms are often linearly decodable in the middle layers, indicating that these quantities are represented explicitly during computation. Although the information encoded in the representations might not be consistent across different instances of the model with separate datasets as contexts, we can still see success in extracting different quantities from a single large context with all the datasets combined. We also observe that the correct answer emerges earlier in the forward pass than the final prediction head, showing that the model converges on the correct value before the last layer projects it into the output space. These findings suggest that TabFMs perform multi step structured computation internally, rather than relying on an opaque end to end mapping.
Our contributions are as follows.

\begin{enumerate}
    \item We present the first mechanistic and representational analysis of TabPFN models, to the best of our knowledge.
    \item We show that both linear coefficients and intermediate arithmetic quantities are embedded in hidden states in a linearly decodable manner.
    \item We trace how the model forms the final answer and characterize how this signal evolves through the depth of the transformer.
    \item We provide empirical evidence that TabFMs implement structured internal computation, pointing toward the possibility of more interpretable and trustworthy tabular foundation models.
\end{enumerate}

\section{Related Work}
While tree-based methods like XGBoost have historically dominated tabular data tasks \citep{grinsztajn2022tree}, recent work has shifted toward Deep Learning architectures designed for tabular modalities \citep{gorishniy2021revisiting}. A significant paradigm shift was the introduction of Prior-Data Fitted Networks (PFNs), most notably TabPFN \citep{hollmann2023tabpfn}. Unlike traditional transfer learning, TabPFN is a Transformer pre-trained on a vast corpus of synthetic datasets generated from Structural Causal Models (SCMs) \citep{causalinferencelearningbook,pearl_2009}. It acts as a proxy for Bayesian inference, solving new classification tasks in a single forward pass via In-Context Learning (ICL) \citep{ye2025closerlooktabpfnv2}. Subsequent iterations, such as TabPFN-v2, have addressed scalability limitations regarding context length and feature counts, with a new improved architecture \citep{hollmann2025tabpfn}.However, interpretability for these models has largely remained restricted to post-hoc feature importance methods like Shapley Value's or Representation Analysis and other classical techniques \citep{ye2025closerlooktabpfnv2,hollmann2025tabpfn,Rundel_2024} rather than mechanistic investigation.

As TabPFN relies on ICL to approximate classification functions, understanding the mechanics of ICL is crucial. In Large Language Models (LLMs), ICL is hypothesized to emerge from specific attention circuits. \citet{olsson2022incontextlearninginductionheads} identified induction heads which are circuits that implement a "copying" mechanism by attending to previous tokens that appeared in similar contexts as a primary driver of ICL.  Furthermore, \citet{hendel2023incontextlearningcreatestask} and \citet{todd2024functionvectorslargelanguage} introduced the concept of Task Vectors (or Function Vectors), showing that the "task" specified by the context (e.g., a classification rule) is compressed into a distinct, manipulable direction within the model's residual stream.

Our work is grounded in the methodology of linear probing \citep{alain2018understandingintermediatelayersusing} seeking to decode information stored in intermediate layers. While this is well-established in NLP \citep{belinkov2021probingclassifierspromisesshortcomings}, recent work has extended these techniques to continuous-value domains closer to tabular data. \citet{wiliński2025exploringrepresentationsinterventionstime} analyzed representations in Time Series Foundation Models (TSFMs), employing Centered Kernel Alignment (CKA) to map layer-wise similarity and activation steering to identify steerable concepts (e.g., trend and seasonality) within the latent space. They demonstrated that, much like LLMs, continuous-domain Transformers learn structured, disentangled representations of high-level concepts. We extend this line of inquiry to Tabular FMs, investigating whether TabPFN similarly encodes decision boundaries and feature interactions in its residual stream. To the best of our knowledge, this represents the first mechanistic interpretability study of TabPFN models.

\section{Experiments and Results}

\subsection{Preliminaries} 
Transformer models have been used extensively in both NLP and computer vision to capture relationships within sequential inputs. This paradigm has been extended to structured tabular data, where transformer-based approaches have become competitive with classical models. A tabular dataset is defined consisting of data in which each row $x_i$ consists of $d$ features or attributes, where $d$ typically varies across datasets with a label $y_i$ belongs to $[C] = \{1, \ldots, C\}$ for a classification task or is a continuous numerical value for a regression task.

 Among transformer models for tabular learning, TabPFN currently represents the state of the art, with widespread adoption in both industry and academia. TabPFN performs in-context learning to predict the test set given the train set, without requiring further parameter updates. To enable this it is pretrained on a large collection of synthetic datasets and learns to approximate Bayesian inference over this prior distribution.

 The primary model used in this paper is TabPFN v2, thus necessitating a deeper dive into its architecture.
\paragraph{TabPFN v2.}
The first step of the pre-processing pipeline includes numerical encoding of categorical features and normalisation of all the features.
The model then embeds the $d$ features into a $k$-dimensional space, adding perturbations to differentiate the features.
Alongside the label embedding $\tilde{y}_i \in \mathbb{R}^k$, each training instance $x_i$ is represented by $(d+1)$ tokens with dimension $k$.  
For a test instance $x^*$, a dummy label is used to generate the label embedding $\tilde{y}^*$. 
The full shape of the input is thus represented as a tensor of shape \((N + 1) \times (d + 1) \times k\).
TabPFN v2 uses two types of self-attention, one after the other, the first being over samples known as sample attention and the other over attributes known as feature attention, which combine to enable in-context learning across either axis.
At the end of the transformer layers the output token corresponding to the dummy label $\tilde{y}^*$ is extracted and mapped to a 10-class logit for classification or is mapped to a single number in case of regression. 

\subsection{Probing for coefficients}
\label{probing_coefficient}

\begin{figure}[h!]
    \centering
    \includegraphics[width=\columnwidth]{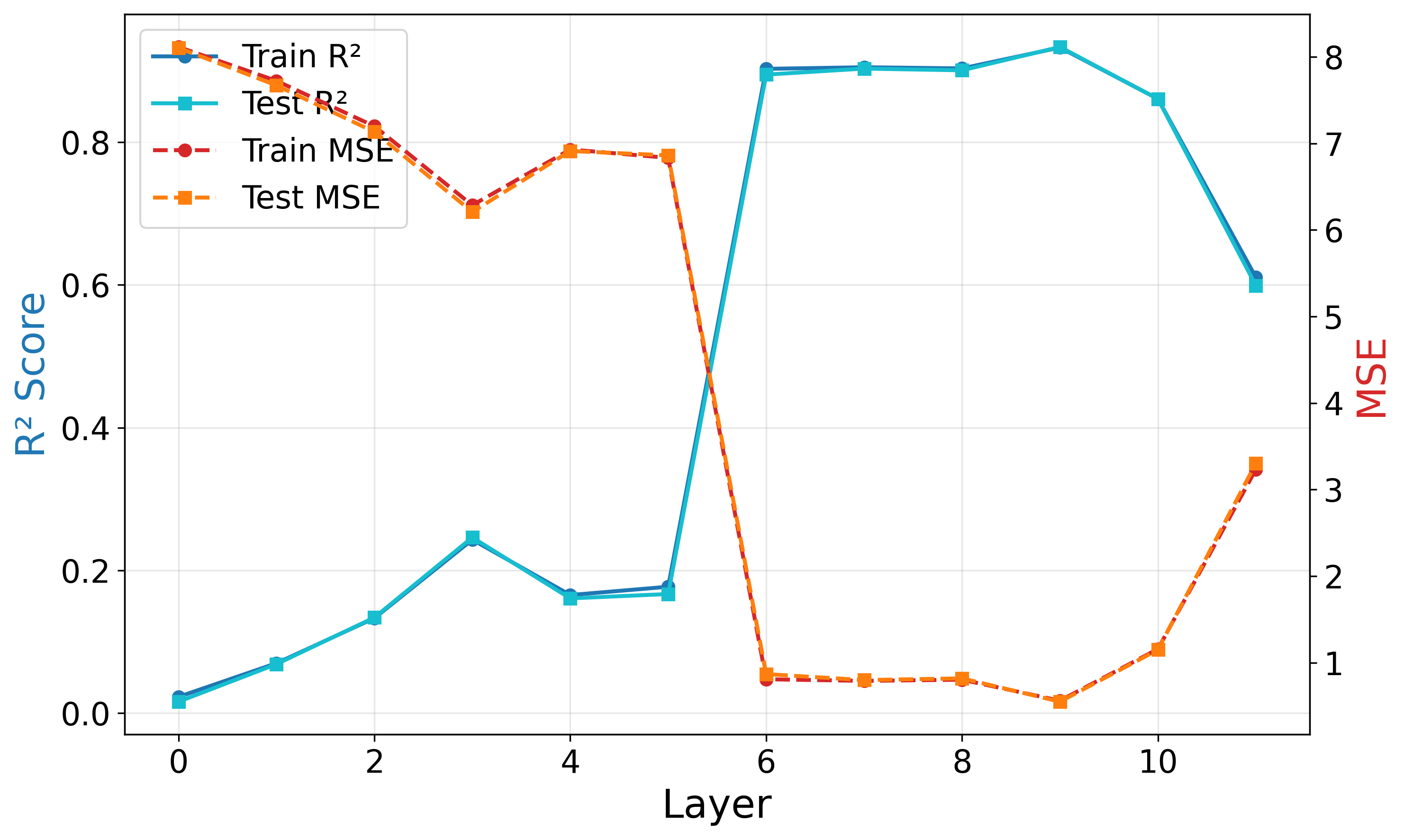}
    \vspace{-6pt}
    \caption{\textbf{Coefficient probe across layers}. Probing $R^2$ and $\mathrm{MSE}$ for coefficient of the linear relationship plotted across different layer activations. The probe $R^2$ sees a sharp increase at layer 6 and drops off at the last layer accompanied by the inverse behaviour in the $\mathrm{MSE}$. High $R^2$ and low $\mathrm{MSE}$ values indicate a better performing probe.}
    \label{fig:coefficient_layers}

    \hfill
    
    \includegraphics[width=\columnwidth]{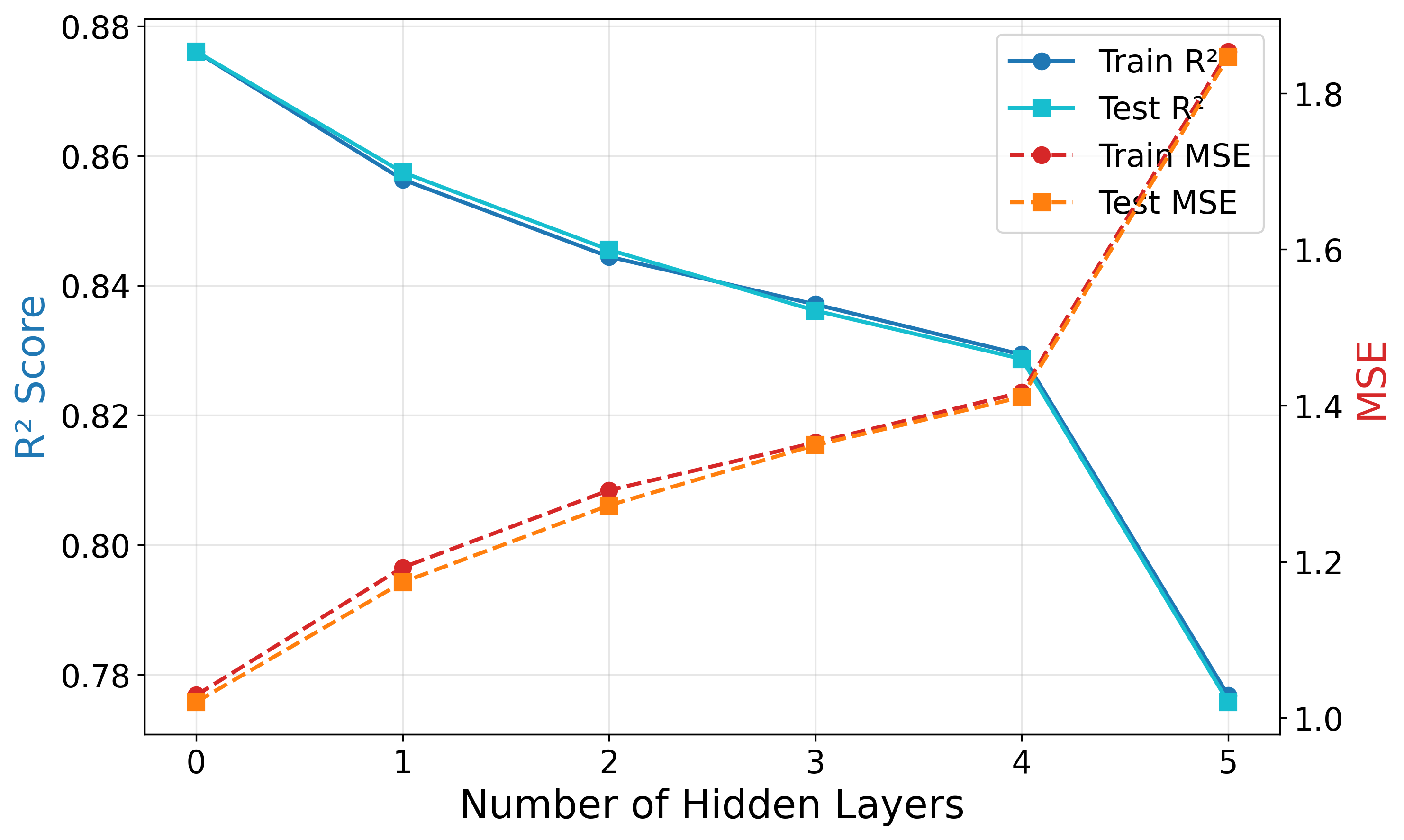}
    \vspace{-6pt}
    \caption{\textbf{Coefficient probe across complexities}. Probing $R^2$ and $\mathrm{MSE}$ for coefficient of the linear relationship plotted across increasing probe complexities, adding hidden layers to an MLP. The probe $R^2$ decreases and the $\mathrm{MSE}$ increases with increasing probe complexity, giving evidence of a linear encoding of the coefficients within the activations.}
    \label{fig:coefficient_complexity}

    \vspace{-1em}
\end{figure}

As TabPFN is an In-Context Learning based model, previous work has shown that coefficients of linear relationships can be extracted from transformers when they are given train dataset as context and asked to predict the answer for test samples. Building on this we investigate whether TabPFN's internal representations encode the coefficients of the modeled linear function $$z = \alpha x + \beta y$$.
To recover these coefficients from the activations, we use two distinct probing setups.
A probe is a function
\[
\phi : \mathbb{R}^{d} \to \mathcal{Y},
\]
which maps an activation tensor
\(
A \in \mathbb{R}^{d}
\)
to a target property, in this case
\(
\alpha \in \mathcal{Y}.
\)
Probes are typically chosen to be simple, lightweight functions, such as linear maps or a shallow neural network.

\textbf{First configuration}. We vary $\alpha$ and $\beta$ across multiple datasets and then fit separate instances of TabPFN for each of these datasets. We can write dataset $n$ thus as:
\[
f_n(x) = \alpha_n x + \beta_n x
\]
We then extract forat a single layer for all datapoints activation $A \in \mathbb{R}^{n \times d \times k}$ and flatten it to get a activation $A \in \mathbb{R}^{ndk}$ vector, trying to learn the probe. We collate these activations for all the different datasets to construct the probing dataset.

The probing dataset is a mapping from this activation vector to the coefficient $\alpha$ or $\beta$.
\[
\phi : \mathbb{R}^{ndk} \to (\alpha, \beta)
\]
This approach yields poor training accuracy. The low probe performance indicates that each TabPFN fit encodes the coefficients differently, so when activations from many fits are pooled, the coefficients no longer align in any common representation that a probe can read out.

\textbf{Second configuration}. In this setup we combine the multiple datasets with varying $\alpha$ and $\beta$ into one large dataset with another added input variable which is unique and maps directly to the different coefficient combinations. So, this added variable acts like a switch input that switches between the coefficient pairs.
\[
f(x; u) = \alpha(u)\, x + \beta(u)\, x
\]
With this dataset we fit the TabPFN model with dataset containing different coefficient pairs and then extract activations from a holdout test set. From these extracted activations we construct the probing dataset which is a mapping from the activations of all the tokens but the switch token to the coefficients. The individual activations which are of shape $A \in \mathbb{R}^{d \times k}$ are flattened to $A \in \mathbb{R}^{dk}$.
The probing dataset thus is a mapping from the activations of a single data point to the coefficient $\alpha$ or $\beta$.
\[
\phi : \mathbb{R}^{dk} \to (\alpha, \beta)
\]

 The results for this setup, as show in \hyperref[fig:coefficient_layers]{Figure 1}, indicate high train and test performance for the probes, showing that we are able to extract the coefficients from the internal activations from a single fit.

This performance disparity between the first and the second configuration indicates that the successful probes in the second configuration setting likely rely on distinguishing between dataset-specific signatures in the coefficients of the linear expressions rather than decoding the coefficient themselves. We posit from this evidence that the coefficients are not encoded as a universal quantity in the representation as shown by the low training accuracy on different fits, but it can be extracted from the representations to a degree as shown by the same fit probing results.

We also run the above probing experiments with increasingly complex probes. The complexity of the probes is controlled by adding more hidden layers into the MLP. The results in \hyperref[fig:coefficient_complexity]{Figure 2} show that with increasing complexity, the $R^2$ scores decrease. This gives evidence for the linear encoding of the coefficients within the activations of the model.

\subsection{Probing for Intermediaries}
\label{sec:intermediary_probing}

\begin{figure}[h!]
    \centering
    \includegraphics[width=\columnwidth]{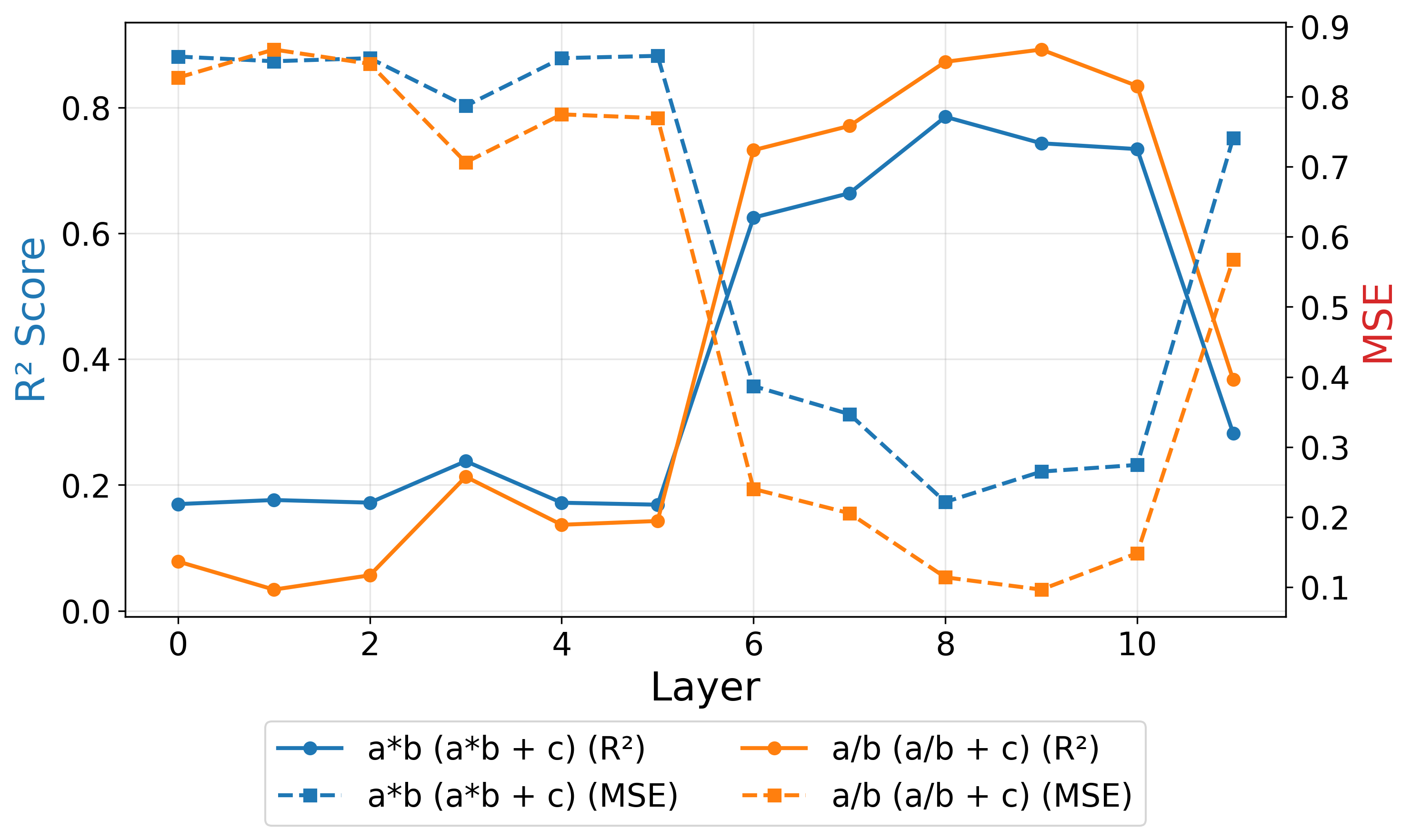}
    \vspace{-6pt}
    \caption{\textbf{Intermediary probe across layers}. Probing $R^2$ and $\mathrm{MSE}$ for the intermediary of the expression plotted across different layer activations. Two different expressions and intermediaries displayed. The probe $R^2$ sees a sharp increase at layer 6 and drops off at the last layer accompanied by the inverse behaviour in the $\mathrm{MSE}$. High $R^2$ values and low $\mathrm{MSE}$ indicate a better performing probe.}
    \label{fig:intermediate_layers}

    \hfill
    
    \includegraphics[width=\columnwidth]{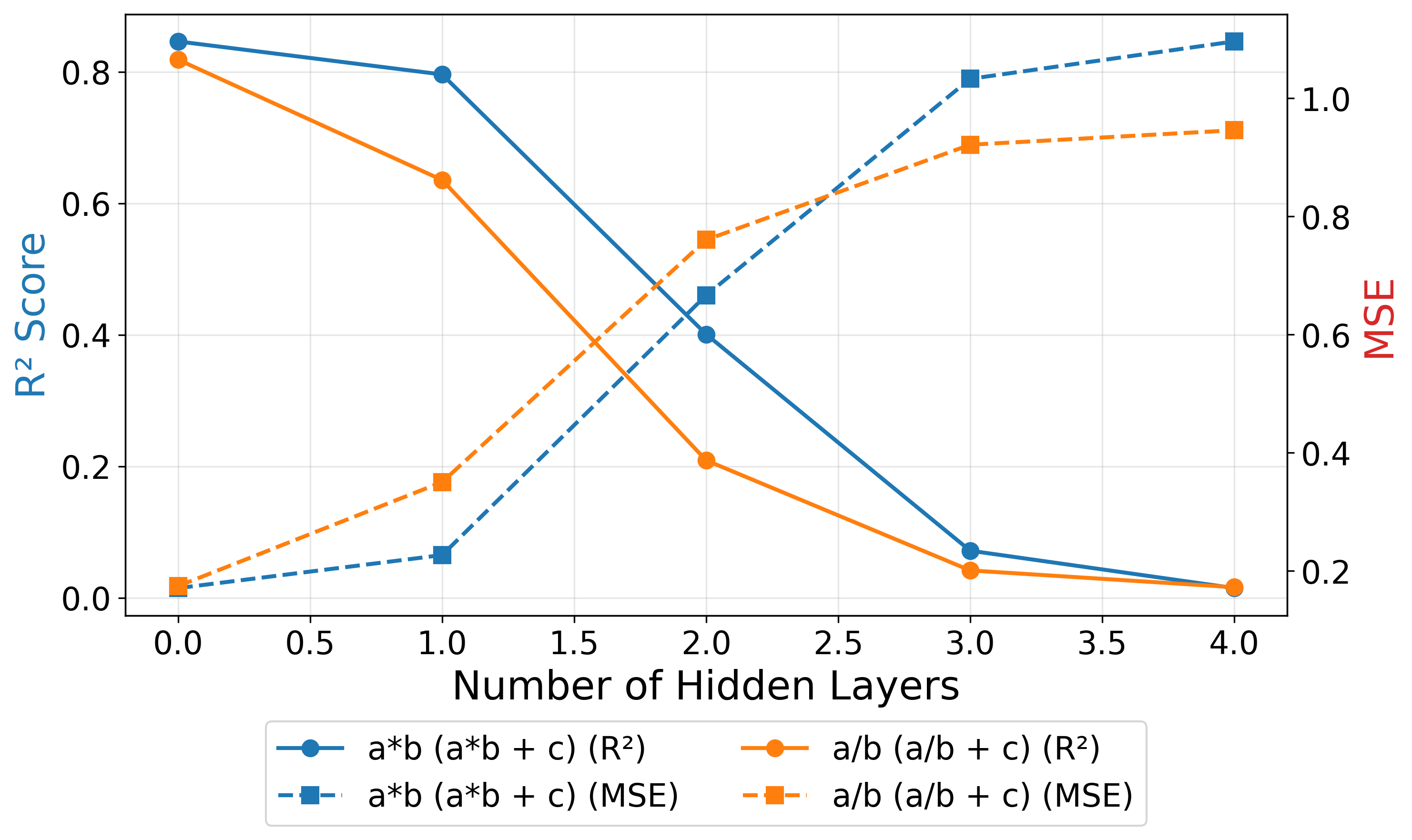}
    \vspace{-6pt}
    \caption{\textbf{Intermediary probe across complexities}. Probing $R^2$ and $\mathrm{MSE}$ for the intermediary of the expression plotted across increasing probe complexities, adding hidden layers to an MLP. The probe $R^2$ decreases with increasing probe complexity, giving evidence of linear encoding of the intermediary within the activations.}
    \label{fig:intermediate_complexity}

\end{figure}

To further investigate the model's internal computation, we formulate an experiment to probe for intermediate expressions required to compute the final answer. This might reveal if the model is working through the expression in a hierarchical manner, computing simpler products and then combining to get the final answer.

So, to elucidate the compositional nature of TabPFN's internal reasoning, we constructed a dataset of an arithmetic relationship $$z = a \cdot b + c$$ In this setting, the product $a \cdot b$ acts as a necessary intermediate variable that must be computed before the final addition with $c$. We aim to probe for this intermediate product

We fit a TabPFN instance on this dataset and then extract the internal activations of the test set during the forward pass. We then construct the probing dataset by mapping the internal activation to the intermediate product for that sample.
\[
\phi : \mathbb{R}^{dk} \to a \cdot b
\]
We train a linear probe on this dataset which shows that the representation of $a \cdot b$ is highly recoverable in the middle layers of the model, before fading in later layers as the final output $z$ is constructed. This localization of the intermediate term provides evidence that TabPFN performs structured, hierarchical computation, effectively mirroring the mathematical order of operations within its internal activations.

Again, as explained in section \ref{sec:intermediary_probing} we repeat the probing experiments for increasingly complex probes. The results in Figure \ref{fig:coefficient_complexity} show that with increasing complexity, the $R^2$ scores decrease. This gives evidence for the linear encoding of the intermediaries in the internal representations of the model.

\subsection{Probing for the answer and Logit Lens}
\label{subsec:answer_probe}
\begin{figure}[h!]
    \centering
    \includegraphics[width=\columnwidth]{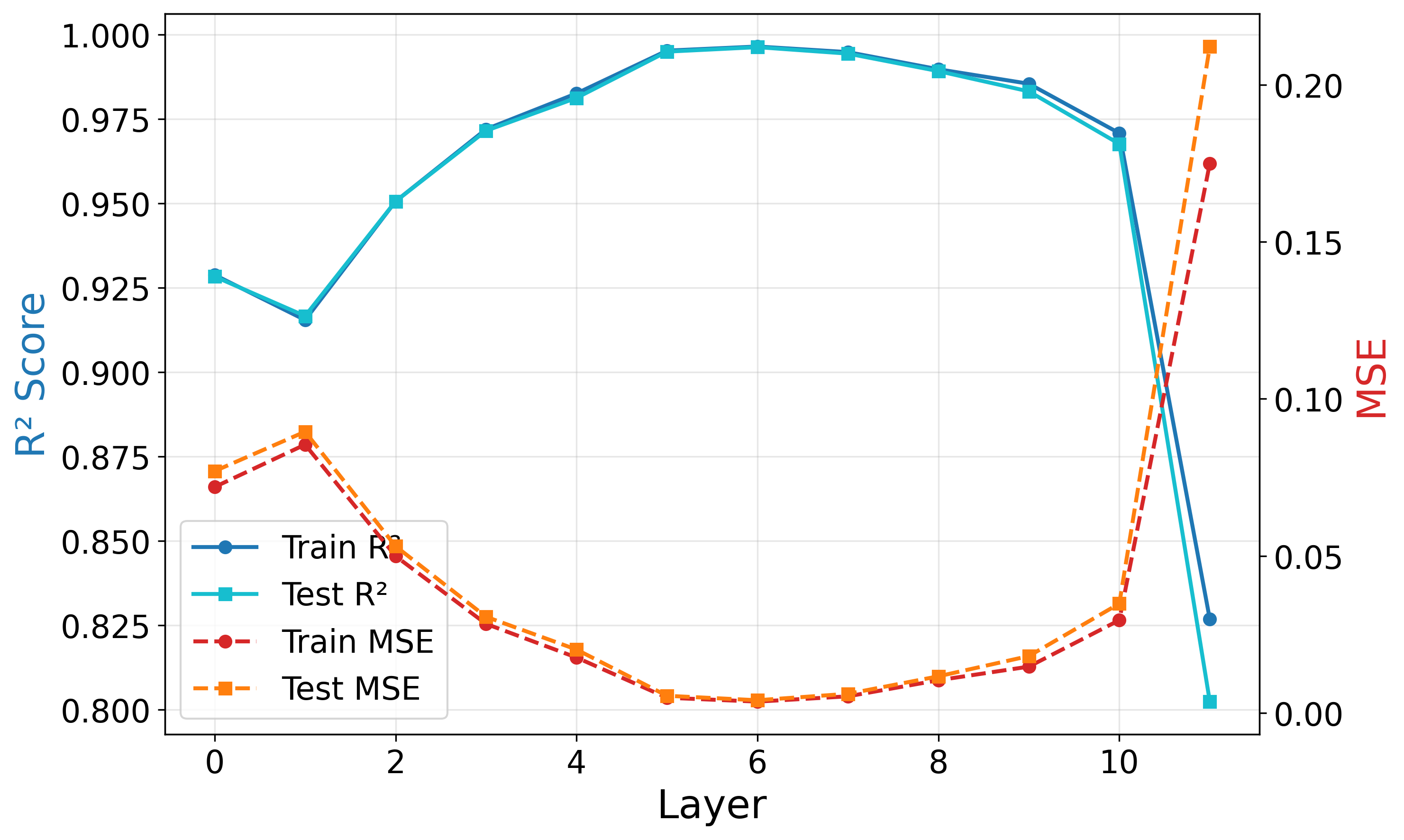}
    \caption{\textbf{Answer probe across layers}. Probing $R^2$ and $\mathrm{MSE}$ for the answer of the expression in the answer token activations plotted across different layer activations. The probe reaches a very high $R^2$ at layer 5 and stays consistent, till an expected drop at the last layer accompanied by the inverse behaviour in the $\mathrm{MSE}$. This indicates the probes effectiveness at extracting the answer earlier in the model through its activations.}
    \label{fig:answer_layers}

    \hfill
    
    \includegraphics[width=\columnwidth]{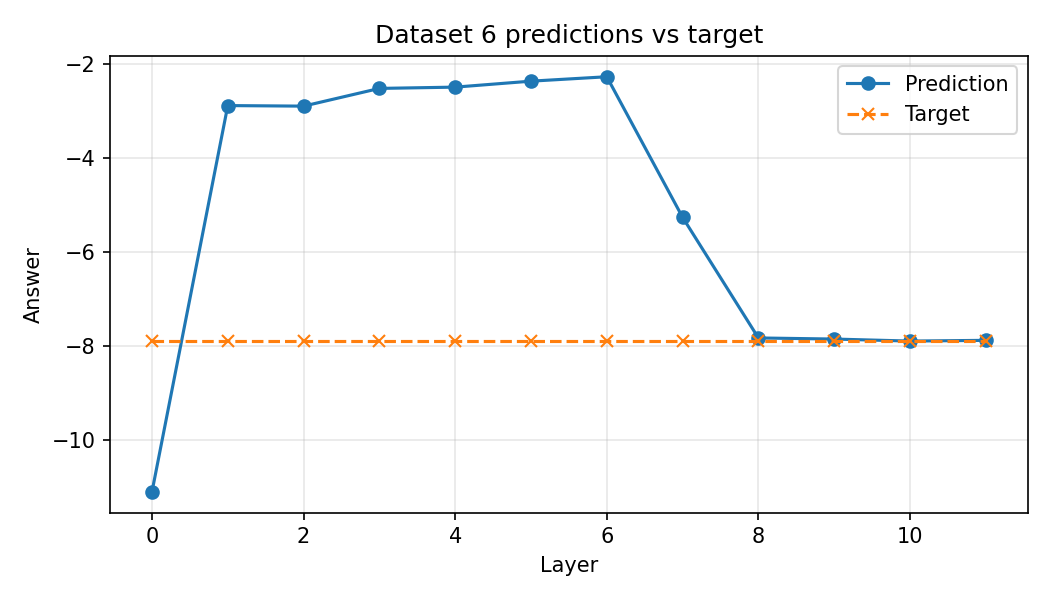}
    \vspace{-15pt}
    \caption{\textbf{Logit Lens results}. This shows the results of the logit lens plotted against the model layer where it is applied. Here, we can see the the result converges to the true answer around layer 8.}
    \label{fig:logit_lens}

\end{figure}

We now moved to experiments aimed at localizing where the answer is formed withing the model's computation. Taking from previous work on LLM interpretability \citep{zhao2023explainabilitylargelanguagemodels}, to investigate how the solution emerges across the network depth, we employed the logit lens technique \citep{nostalgebraist2020interpreting}. In this technique we apply the model's final unembedding matrix directly to the intermediate layer activations aiming to find how the answer is refined through the layers.

\[
\operatorname{LogitLens}_L :
h_L(x) \mapsto W_U \, h_L(x)
\]

We use the same setup as \hyperref[probing_coefficient]{Section 3.2} to fit TabPFN models on linear relationships. We observed that the internal representations strictly align with the valid output space only in the later stages of the network, with Layer 7 proving critical for this alignment in the linear regression task.

However, extracting the output using trained linear probes reveals a contrasting dynamic. For relatively simple arithmetic relationships both linear regression and the compound $z = a + b \cdot c$ task, we found that the final answer $z$ can be linearly decoded with high effectiveness as early as Layer 5. The discrepancy between the probe-accessible solution (Layer 5) and the native output alignment (Layer 8) suggests a degree of computational inefficiency for simple tasks. TabPFN appears to compute the solution early but continues to process the representation through several subsequent layers before projecting it onto the final output manifold much later.

\subsection{Probing for inputs in answer token}

\begin{figure}[h!]
    \centering
    \includegraphics[width=\columnwidth]{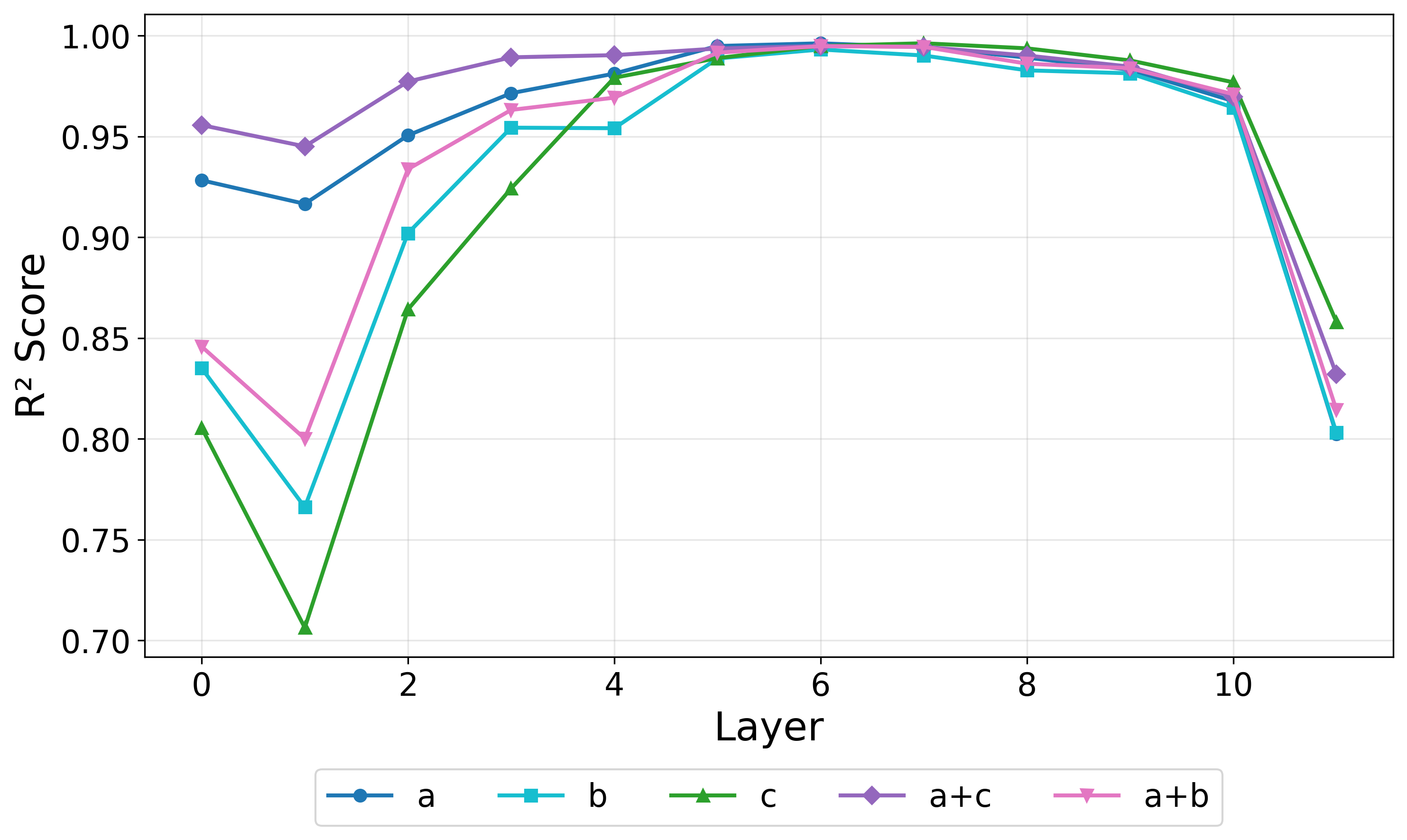}
    \caption{\textbf{Input probe across layers}. Probing $R^2$ for the inputs and their linear combination in the answer token activations plotted across different layer activations. The probe reaches a very high $R^2$ at layer 5 and stays consistent, till an expected drop at the last layer. This indicates the probes effectiveness at extracting the inputs purely from the answer tokens indicating some kind of copying behavior in the earlier parts of the model.}
    \label{fig:input_probe_results}

\end{figure}

Building on earlier work in large language model interpretability, which identified copy and induction heads that move information between tokens for later computation, we design an experiment to probe for input information directly in the answer token. Similiar to \ref{subsec:answer_probe}, we create a dataset with the compound expression $$z = a \cdot b + c$$ 
Our goal is to test whether the input values and their linear combinations appear in the activations of the answer token. We extract the activations of the TabPFN test set only at the last index, which corresponds to the asnwer token, $y \in \mathbb{R}^d$.Using these activations, we form a probing dataset by mapping them to the different input values and their combinations.
\[
\phi_y : \mathbb{R}^{d} \to (a \cdot b,a,b,c)
\]
Figure \ref{fig:input_probe_results} shows that these quantities can be recovered early in the model. This suggests that the model copies input information forward and encodes it in different directions within the answer token subspace.

\section{Discussion and Future Work}
In this work, we presented a mechanistic analysis of TabPFN, moving beyond post-hoc feature importance to investigate the internal algorithmic structure of Tabular Foundational Models. Our experiments shine light on the TabPFN black-box and reveal that TabPFN constructs its answers systematically across several layers with varying functions and importance, with intermediate results linearly represented in the residual stream.

Our results regarding intermediate value probing offer strong evidence that TabPFN implements systematic reasoning. By successfully recovering the intermediate term $a \cdot b$ in the calculation of $z = a \cdot b + c$, particularly in the middle layers, we show that the model systematically constructs the answer, respecting the mathematical order of operations within its residual stream.
Probing for coefficients revealed that TabPFN may not generalize across fits but does generalize within them, i.e similar functions may be represented differently across fits, but have same or similar representation within the same fit, this is demonstrated by the failure of probes to generalize across different fits in section \ref{probing_coefficient} (first configuration), contrasted with their high success within a single fit in section \ref{probing_coefficient} (second configuration). These results imply that TabPFN does not maintain fixed, universal encoding of the coefficients in its internal representations.

The results in the probing and logit lens section again indicate that although the semantic content of the solution is computed early, the subsequent layers are dedicated to transforming this representation into the specific geometric alignment required by the output head with certain layers, such as layer 7, being important for this shift in representation.

Our analysis is currently limited to simple arithmetic (additive/multiplicative) toy expressions compared to the highly complex noisy functions found in real world datasets which can be improved upon in future work.
Future work could also entail finding the mechanism/mechanisms through which the answer is computed using mechanistic interpretability methods like activation patching, or trying to manipulate the function through steering vectors or vector ablation.

\bibliography{example_paper}
\bibliographystyle{icml2025}

\newpage



\end{document}